\documentclass[final]{cvpr}
\usepackage{times}
\usepackage{epsfig}
\usepackage{graphicx}
\usepackage{amsmath}
\usepackage{amssymb}
\usepackage{placeins}
\usepackage{booktabs}
\usepackage[misc]{ifsym}

\DeclareMathOperator*{\argmin}{arg\,min}

\usepackage[pagebackref=true,breaklinks=true,colorlinks,bookmarks=false]{hyperref}

\def\cvprPaperID{5} 
\def\confYear{CVPR 2021}
\begin{document}
%
\title{Towards Automated and Marker-less Parkinson Disease Assessment: Predicting UPDRS Scores using Sit-stand videos}
\author{Deval Mehta\textsuperscript{1}$^{\textrm{\Letter}}$, Umar Asif\textsuperscript{1}$^{\textrm{\Letter}}$, Tian Hao\textsuperscript{2}$^{\textrm{\Letter}}$, Erhan Bilal\textsuperscript{2},\\
Stefan Von Cavallar\textsuperscript{1}, Stefan Harrer\textsuperscript{1,*}, and Jeffrey Rogers\textsuperscript{2}\\ 
\textsuperscript{1}IBM Research, Australia, \textsuperscript{2}IBM Research, Yorktown Heights, New York, USA\\
{\tt\small \{deval.mehta1092,umar.asif\}@gmail.com, \{thao,ebilal,jeffrogers\}@us.ibm.com,}\\
{\tt\small svcavallar@hotmail.com, stefan.harrer@dhcrc.com}
}
\maketitle

\makeatletter
\def\blfootnote{\xdef\@thefnmark{}\@footnotetext}
\makeatother

\begin{abstract}

This paper presents a novel deep learning enabled, video based analysis framework for assessing the Unified Parkinson’s Disease Rating Scale (UPDRS) that can be used in the clinic or at home. We report results from comparing the performance of the framework to that of trained clinicians on a population of 32 Parkinson’s disease (PD) patients. In-person clinical assessments by trained neurologists are used as the ground truth for training our framework and for comparing the performance. We find that the standard sit-to-stand activity can be used to evaluate the UPDRS sub-scores of bradykinesia (BRADY) and posture instability and gait disorders (PIGD). For BRADY we find F1-scores of 0.75 using our framework compared to 0.50 for the video based rater clinicians, while for PIGD we find 0.78 for the framework and 0.45 for the video based rater clinicians. We believe our proposed framework has potential to provide clinically acceptable end points of PD in greater granularity without imposing burdens on patients and clinicians, which empowers a variety of use cases such as passive tracking of PD progression in spaces such as nursing homes, in-home self-assessment, and enhanced tele-medicine.
\end{abstract}

\section{Introduction}
Parkinson’s disease (PD) is a progressive neuro-degenerative disorder affecting 10 million people worldwide with approximately 60,000 Americans being diagnosed each year \cite{understand_pd}. It has been reported that the lifetime risk of developing PD is 2\% for men and 1.3\% for women over the age of 40 \cite{ascherio2016epidemiology}. The hallmarks of PD include motor symptoms such as bradykinesia (\textbf{BRADY}), rigidity, tremor, posture instability and gait disorders (\textbf{PIGD}), as well as non-motor symptoms such as olfactory dysfunction and sleep disorder, imposing profound impacts on the quality of life of PD patients.

\blfootnote{*Now at Digital Health Cooperative Research Centre, Australia.}

There is no cure for PD; however, dopamine replacement therapy can help to manage symptoms with the goal of improving the quality of life as much as possible. A crucial part to ensuring the effectiveness of the treatment is customizing the types, timing and dosage of medications as the disease progresses. It is thus essential for neurologists to understand the severity of symptoms and the extent of motor fluctuations. The current clinical instrument for assessing disease progression and response to treatment is a structured in-person assessment based on protocols specified in Unified Parkinson’s Disease Rating Scale (UPDRS). Specifically, such assessment is administered by trained clinicians once or twice a year with major drawbacks including ``white coat effect", relatively high levels of disagreement in ratings among different neurologists, as well as its lack of ability to characterize the fluctuation of symptoms \cite{post2005unified}.

In an effort to better understand the dynamics of motor symptoms in between clinic assessments, daily self-reports of symptoms \cite{hauser2004parkinson} have been used as an additional tool for clinicians to get a sense of the response to treatment. However, similar to most questionnaire-based approaches, this method suffers from limitations such as recall bias and poor adherence which result in poor accuracy and reliability. Such limitations and the importance of understanding motor fluctuations to PD management have been driving research efforts in deriving continual metrics of PD symptoms from signals using wearable devices \cite{erb2020mhealth, abrami2020using, mahadevan2020development}. The wearable-based approach, although able to provide objectively quantifiable measurements of motor fluctuations, still requires effort from the patients on a daily basis and therefore is potentially susceptible to compliance adherence.

Encouraged by the exciting progress in computer vision and deep learning, we set out to investigate the feasibility of a video-based approach for free-living assessment of PD in a contact-free and passive manner. In this paper, we showcase the idea by developing and evaluating an end-to-end deep learning framework that takes individual video clips containing sit-stand motion as input and outputs predictions of two UPDRS sub-scores, Bradykinesia (BRADY) and PIGD scores, which are commonly used as clinical endpoints of PD. 

The dataset used to support this exploration was collected as part of a clinical study \cite{erb2020mhealth} where video recordings of patients' UPDRS examinations were obtained, along with in-clinic ratings from the neurologist who administered the UPDRS, as well as ratings from two additional clinicians who scored the patients by watching the video recordings of their UPDRS sessions. 

\begin{figure*}[t]
	\begin{center}
		\includegraphics[trim=0.1cm 0.2cm 0.1cm 0.3cm,clip,width=0.86\linewidth,keepaspectratio]{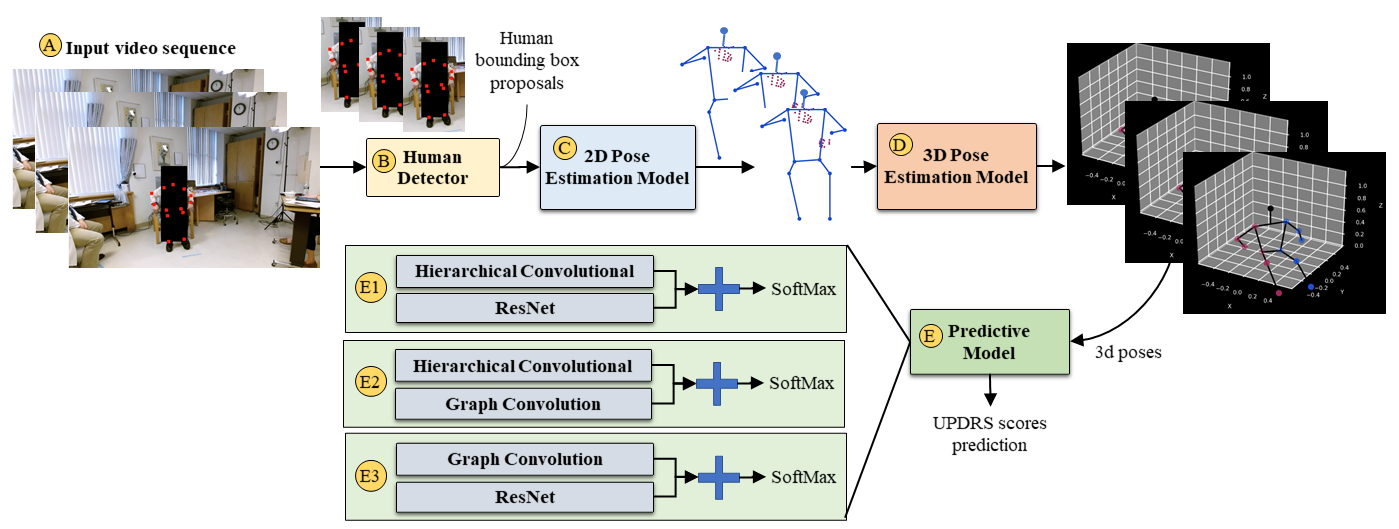}
		\vspace{0pt}
		\caption{Overview of the reported analysis framework. Given an input video sequence (A), first a human detector (B) extracts human proposals from the video frame-by-frame and feed the proposals into a 2D pose estimation model (C) which predicts coordinate locations of human  joints in 2D image space. Next, a 3D pose model (D) uses the 2D pose information and predicts joints locations in 3D Cartesian space. Finally, the 3D pose information is fed into the proposed ensemble zoo (E) to predict UPDRS scores. (E1), (E2), and (E3) are three different ensemble combinations for our predictive models. Note that the input image (A) is overlaid with a black rectangle for privacy purposes.}
		\label{fig_framework}
	\end{center}
	\vspace{-20pt}
\end{figure*}

As shown in Figure \ref{fig_framework}, the proposed framework comprises a sequence of deep learning enabled components that take a short video containing sit-stand motion as input and output predictions of UPDRS subscores including BRADY and PIGD. We believe the proposed technology has potential to transform the way we understand motor symptoms by providing automated and continual predictions of motor examination ratings (UPDRS-III) commonly used as clinical endpoints for PD. We envision the proposed system would enable clinical decision makings and accelerate clinical trials by providing clinical-grade assessments on a daily basis compared to a semi-yearly or yearly basis in today's practice. In addition to its contribution to the delivery and development of treatment, the proposed technology could also provide value to PD patients by offering in-home self-assessment which gives them more control over disease management.

To summarize, our contributions in this paper include the following:
\begin{enumerate}
\item We present a novel analysis framework based on deep learning models which use human body pose data representations to automate assessment of UPDRS sub-scores (i.e., BRADY and PIGD scores) in a contact-free and passive manner while achieving performance comparable to trained neurologists.
\item We present results from retrospective analysis of clinical assessments of PD patients, and compare our analysis framework to trained clinicians evaluating the same video. Our results demonstrate the potential of deep learning enabled ambient sensing to facilitate remote clinical evaluations.
\end{enumerate}

\section{Related Work}
Recently there has been a growing body of research efforts aiming to develop technologies that are able to quantify PD motor fluctuations in free-living conditions. Most of such technologies rely on acceleration data collected from wearable motion sensors to generate relevant features indicative of motor symptoms.

The wearable-based approach can be further divided into the following two categories. Some early works \cite{pastorino2013wearable} used a network of on-body motion sensors to collect acceleration data from various body parts of interest (e.g., wrists, waist, ankles, etc.) to identify kinematic features that are potentially related to motor symptoms. More recently, in an effort to simplify the setup process, and more importantly for the ability to harness data from consumer devices such as smartwatch and fitness trackers, a growing body of research \cite{abrami2020using, mahadevan2020development} has been investigating the feasibility of leveraging only data from the wrist-worn motion sensor to estimating changes in PD state.

In addition to the wearable-based approaches, camera-based methods have been proposed, most of which relied on the 3D skeletal keypoints provided by commercial depth cameras \cite{ms_kinect}. In Galna et al. \cite{galna2014accuracy}, the authors benchmarked depth camera's output against a gold standard motion analysis system to study its accuracy in measuring clinically relevant movements in people with PD. In other studies, kinematic features derived from the depth camera have been used to differentiate subjects with and without PD \cite{eltoukhy2017microsoft}, and to detect freezing of gait \cite{bigy2015recognition}. Compared to wearable-based methods, camera-based approaches offer the advantage of providing more fine-grained spatial-temporal kinematic information and requires less effort for setup due to its contact-free nature. In this paper, we develop a marker-less framework based on RGB cameras which is empowered by recent advances in video-based pose estimation and activity recognition techniques that can automate the process of predicting the UPDRS scores for PD patients.

The rest of the paper is structured as follows. In section 3, we describe the overall framework of our proposed system which includes the tasks from input video processing to prediction of the UPDRS scores. In this section, we also explain the predictive models, their corresponding data representation and our ensemble architecture. In section 4, we provide the details of the dataset along with various visualization of feature embeddings and some samples. In this section, we also outline our train-test protocol and other implementation details. We end this section by showing the results for both our baseline and ensemble models and benchmark our results with those of the clinician video raters. Finally, we conclude this paper in section 5.

\section{Proposed Framework}
Fig \ref{fig_framework} shows the overall architecture of the proposed framework. It starts with a human detector \cite{osokin2018real} which extracts human bounding box proposals from an input video sequence. The human proposals are fed into a 2D pose estimation model \cite{osokin2018real} which predicts coordinate locations of human joints in 2D image space. Next, a 3D pose model \cite{asif2020sshfd} takes the 2D pose as input and predicts joints locations in 3D Cartesian space. Finally, the 3D pose information is fed into an ensemble zoo which contains three different ensemble models predicting the UPDRS scores of the input video sequence. Note that for our experiments, we also consider 2D joints for the predictive models. The example of predictive models shown in Fig \ref{fig_framework} is for our best performing predictive models. In the following, we describe in detail the different baseline models employed, their corresponding data representations and our proposed ensemble for predicting the UPDRS scores.
\subsection{Baseline Networks}
Since our input data is 2D and 3D pose information fetched from the video sequence and the task being to predict the UPDRS scores, we propose to employ the models used in a similar task of skeleton-based action recognition. We start with building a simple baseline model of a 4-layer Multi-layer Perceptron (MLP). Later, we employ the popularly used models for the task of skeleton-based action recognition. Earlier methods for such a task have used Long Short Term Memory Networks (LSTMs) \cite{liu2017skeleton},\cite{si2019attention} and Temporal Convolutional Network (TCN) \cite{kim2017interpretable} for classifying actions based on the skeleton data. We employ these models for our task as well. Later, we experiment with more complex models which have achieved recent state-of-the-art in skeleton based action recognition performance. For this experimental phase, we select the following models - Hierarchical Convolutional Network (HCN) \cite{li2018co}, Spatio-Temporal Graph Convolutional Networks (ST-GCN) \cite{yan2018spatial}, and Convolutional Networks (CNNs) \cite{caetano2019skelemotion} such as Resnet50.
\subsection{Data Representation}
For the above specified networks, we reshape our input data into their accepted formats for each network. For TCN, HCN, and ST-GCN, our data is constructed in the shape of $C \times$ $T \times$ $N_j \times$ $N_p$ representation where $C$ represents the dimensionality of each joint (2 for 2D pose and 3 for 3D pose), $T$ represents the temporal length i.e. the number of frames in a video sample, $N_j$ represents the number of joints which are 13 in our case, $N_p$ represents the number of persons in the video which is 1 in our case. For CNN such as Resnet50, we convert the joint data into visual representation. For this, we reshape the $C \times$ $T \times$ $N_j \times$ $N_p$ data to $C \times$ $T \times$ $N_j$ * $N_p$ and normalize channel values between 0 and 255. Then we resize this visual representation to a fixed representation of ($C\times244\times244$) using bilinear interpolation. For a given joint $j_{t,k}$ of type $k$ at a time frame $t$, the corresponding normalized pixel value is calculated as:
\begin{equation}
    d = 255 \times \frac{j_{t,k} - c_{min}}{c_{max} - c_{min}},
\end{equation}
where $c_{max}$ and $c_{min}$ correspond to the maximum and minimum values of all the joint coordinates in the data respectively. For MLP model, the data shape of $C \times$ $T \times$ $N_j \times$ $N_p$ is reshaped and represented as a wholesome feature with the shape of $C$ * $T$ * $N_j$ * $N_p$. For LSTM model, the data shape of $C \times$ $T \times$ $N_j \times$ $N_p$ is reshaped to $T \times C$ * $N_j$ * $N_p$, in which $T$ corresponds to the sequence length and $C$ * $N_j$ * $N_p$ corresponds to number of features.

\subsection{The Proposed Ensemble}
We also build ensemble architectures to improve the generalization capability of the individual models. For this, we explore different combinations of state-of-the-art HCN, ST-GCN, and Resnet50 models that extract hierarchical, structural and covolutional features from the data, respectively. The individual models are trained independently and combined during inference by summing the final logits produced by the models.
Specifically, each model in the proposed ensemble ends with a global average pooling operation and produces $Y\in \mathbb{R}^{1\times M}-$dimensional feature maps which are then fed to a linear layer of $1\times K$ dimensions to produce probabilitic distributions ($Q_{s}\in\mathbb{R}^{1\times K}$) with respect to $K$ target classes (representing UPDRS scores of BRADY \& PIGD). Mathematically, the output of a linear layer can be written as:
\begin{eqnarray}
Q^{s} = Y*W^s+B^s
\end{eqnarray}
where, $W^s$ and $B^s$ represent weights and bias matrices, respectively. Finally, the outputs of the linear layers are summed to produce a combined feature representation $P_s$. It is given by:
\begin{equation}\label{loss_cls}
P_s=\sum_{i=1}^{N_s}Q^{s}_i
\end{equation}
Consider a training dataset of videos and labels $(x,y)\in (\mathcal{X}, \mathcal{Y})$, where each sample belongs to one of the $K$ classes $(\mathcal{Y}={1,2,...,K})$. 
To learn the mapping $f_{s}(x): \mathcal{X}\rightarrow \mathcal{Y}$, we train our ensemble models parameterized by $f_{s}(x,\theta^*)$, where $\theta^*$ are the learned parameters obtained by minimizing a training objective function $\mathcal{L}_{train}$:
\begin{eqnarray}\label{eq_seg} 
\theta^{*}=\argmin_{\theta}\mathcal{L}_{train}(y,f_{s}(x,\theta))
\end{eqnarray}
Our training function $\mathcal{L}_{train}$ is based on a CrossEntropy loss which is applied on the outputs of the individual models of the ensemble with respect to the ground truth labels ($y$). Mathematically, $\mathcal{L}_{train}$ can be written as:
\begin{eqnarray}\label{eq_seg} 
\mathcal{L}_{train}(Q_{s},y)=\sum_{k=1}^K\mathbb{I}(k=y)\log\sigma (Q_{s},y),
\end{eqnarray}
where $\mathbb{I}$ is the indicator function and $\sigma$ is the SoftMax operation. It is given by: 
\begin{equation}
\sigma(z)=\frac{\exp(z)}{\sum_{k=1}^K \exp(z_k)}. 
\end{equation} 
\section{Experiments}

\subsection{Dataset}
The present study was a non-interventional study conducted at a single site in 35 people with mild (Hoehn \& Yahr = \{1\}) to moderate (Hoehn \& Yahr = \{3\}) PD (Table \ref{table_study}) under controlled laboratory conditions. Participants were asked to perform a series of motor and cognitive tests in accordance with the UPDRS protocol over two visits lasting approximately one hour each. During one visit the study participant was off medication and thus in an "OFF" state, while in the other visit the study participant had recently taken his/hers medication and thus was in an "ON" state. The order of the ON and OFF visits was randomized and no more than 14 days passed between the visits. The sessions were video taped by a technician and led by the same neurologist who also scored each UPDRS task. Later on, the videos were watched and independently scored by two other clinicians.  Data collection was carried out at the clinical and translational research center (CTRC) at Tufts Medical Center and all study procedures were approved by the Tufts Health Sciences Campus Institutional Review Board. More details on the study protocol can be found in Mahadevan et al. \cite{mahadevan2020development}. 

\begin{table}[t!]
	\caption{Summary of subjects in the study (N = 35)}
	\vspace{5pt}
	\label{table_study}
	\centering
	\setlength\tabcolsep{5.0pt}
	\begin{tabular}{@{}ll@{}}
		\hline
		Age (years) & $68.3 \pm  8.0 \: (46 - 79)$\\
		Height (cm) & $171.6 \pm  16.4 \: (147 - 189)$\\
		Weight (kg) & $82.3 \pm  18.0 \: (44 - 112)$\\
		Gender (\%)\\
		\: Male & $23 \: (65.7\%)$\\
		\: Female & $12 \: (34.3\%)$\\
		Hoehn \& Yahr (\%)\\
		\: 1 & $2 \: (6\%)$\\
		\: 2 & $26 \: (74\%)$\\
		\: 3 & $7 \: (20\%)$\\

		\hline

	\end{tabular}
	\vspace{5pt}
\end{table}

\subsection{Class Distribution and Train-test Split}
\begin{figure*}[t]
    \begin{center}
        \includegraphics[trim=0.1cm 0.2cm 0.1cm 0.3cm,clip,width=1.0\linewidth,keepaspectratio]{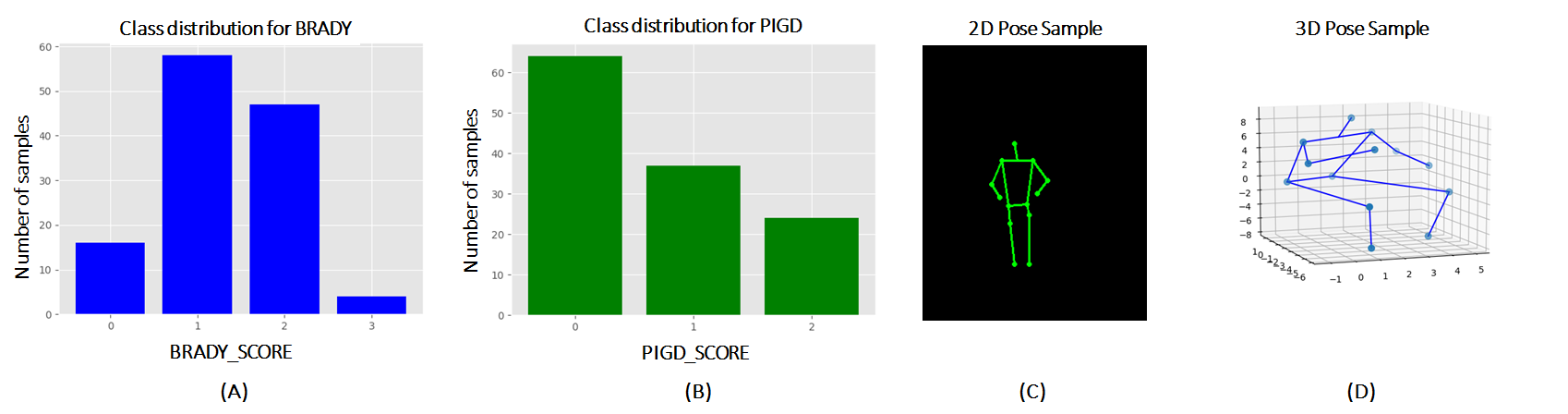}
        \vspace{-12pt}
        \caption{Class distribution for (A) BRADY scores and (B) PIGD scores. Dataset samples for (C) 2D Pose and (D) 3D Pose.}
        \label{fig_class_dist}
    \end{center}
\end{figure*}

We frame the task of predicting the UPDRS scores from the input videos as a classification task by converting the average neurologist rated values of the BRADY and PIGD scores to integers. Based on the data collected, for BRADY, this translates to a 4-class problem \{0, 1, 2, 3\} and for PIGD it translates to a 3-class problem \{0, 1, 2\}, as there were no cases having a PIGD score of \{3\} for this data. With this method, we present the distribution for 125 video samples in our dataset amongst the different classes for both BRADY and PIGD in Fig \ref{fig_class_dist}. Note that only videos with ratings from the on-site neurologist and both clinician video raters were selected, resulting in a dataset involving sit-stand videos from 32 subjects.

We employ a 5-fold train-test split protocol. We use the \textit{stratified fold} split strategy so that the class distribution is preserved amongst the train and test samples. With this protocol, we get 5-folds with each fold containing 100 training sample videos and 25 testing sample videos. When comparing the performance of our trained models with that of the clinician video raters, we evaluate them on each fold and employ weighted averaging for the scores presented later in the Results section.
\begin{figure*}[t]
    \begin{center}
        \includegraphics[trim=0.1cm 0.2cm 0.1cm 0.3cm,clip,width=0.9\linewidth,keepaspectratio]{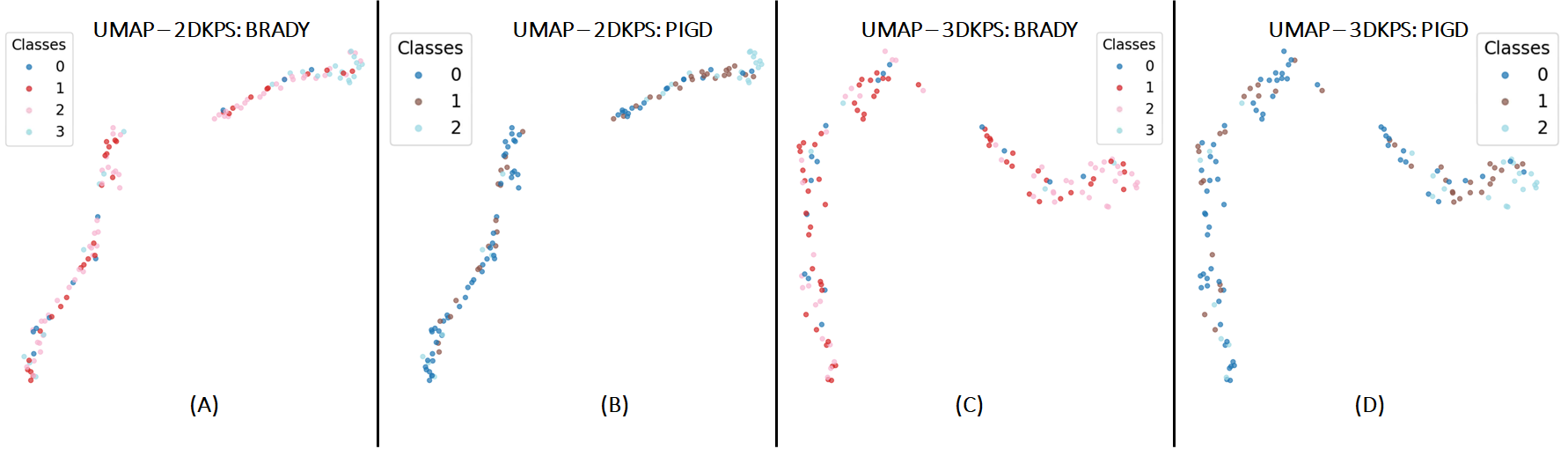}
        \caption{UMAP embeddings clusters with respect to individual classes of (A) BRADY for 2D joints (B) PIGD for 2D joints (C) BRADY for 3D joints (D) PIGD for 3D joints}
        \label{fig_umap_all}
    \end{center}
\end{figure*}
\subsection{Data Distribution and Feature Embeddings}
We also show the distribution of the feature embeddings amongst the different classes based on the body 2D and 3D joints. For this purpose, we apply the dimension reduction technique of Uniform Manifold Approximation and Projection (UMAP) \cite{mcinnes2018umap} to map out the body 2D joints and 3D joints distribution amongst the classes of BRADY and PIGD. Fig \ref{fig_umap_all} shows the distribution of feature map embeddings of the 2D joints and 3D joints with respect to the sample classes in BRADY and PIGD. It can be seen from Fig \ref{fig_umap_all} that the feature embeddings are highly scattered in nature where there are no pre-defined cluster patterns for the different classes both for BRADY and PIGD, which makes the task of predicting the UPDRS scores difficult.
\subsection{Implementation Details}
We trained all our models both baseline and ensembles using Adam optimizer with the default parameters except having the weight decay of 1e-4, batch size of 16, an initial learning rate of 0.001 with an exponential decay factor of 0.99 for 100 epochs. Both the sequential 2D and 3D joints are transformed to a fixed temporal length width of \textit{T} = 150 frames with bilinear interpolation along the temporal dimension. \textit{T} = 150 is selected as the mean length of the temporal lengths of all the video samples in our dataset.

As we are converting the raw BRADY \& PIGD scores to integers, this can lead to difficulty for our models in learning some closely related features. Moreover, for a better generalization of the models and to avoid over-fitting on our classification tasks, we implement the training with \textit{label smoothing} strategy \cite{pereyra2017regularizing},\cite{muller2019does}. With this strategy, we minimize the cross-entropy loss between the modified targets $y^{LS}$ and the network's output probability. The modified target $y^{LS}$ is given as:
\begin{equation}\label{label_smooth}
y^{LS} = y(1-\alpha) + \alpha/K
\end{equation}
where $\alpha$ is the label smoothing parameter, \textit{y} is the original target of a sample and \textit{K} represents the number of categorical classes for the task. For our experiments, we vary the parameter $\alpha$ in the range [0.05, 0.1, 0.2, 0.3, 0.4] and found that 0.2 produces the best results.
\begin{figure*}[t]
    \begin{center}
        \includegraphics[trim=0.1cm 0.2cm 0.1cm 0.3cm,clip,width=1.0\linewidth,keepaspectratio]{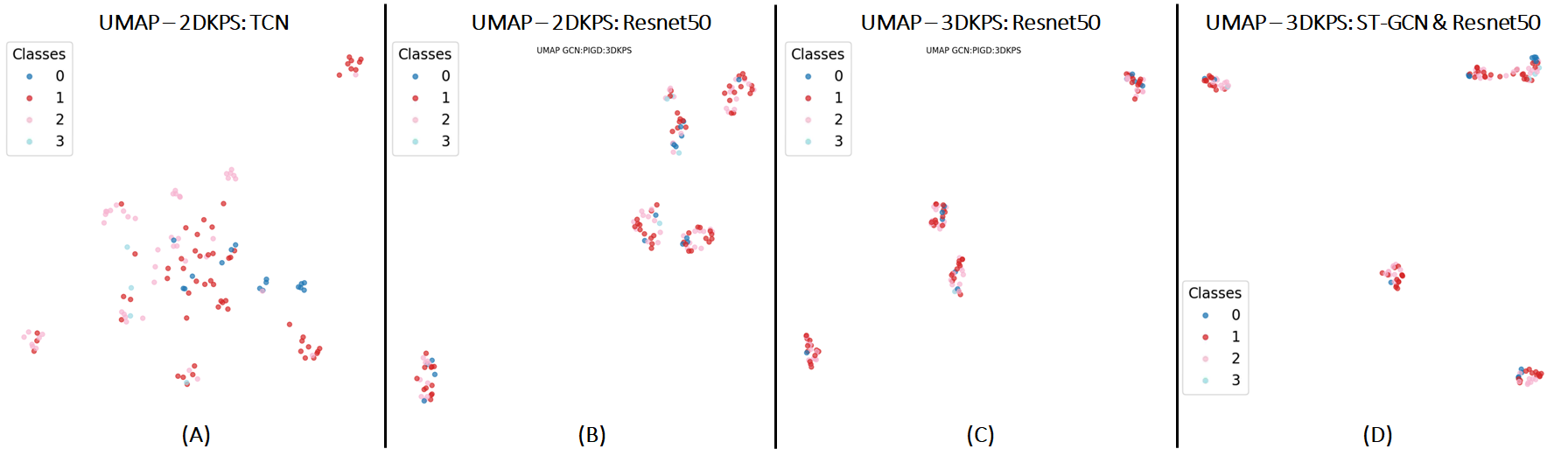}
        \caption{UMAP embedding clusters based on the representative features for BRADY classification on test data for different models and input data (A) TCN based on 2D joints (B) Resnet50 based on 2D joints (C) Resnet50 based on 3D joints (D) Ensemble of ST-GCN + Resnet50 based on 3D joints}
        \label{fig_umap_brady_net}
    \end{center}
\end{figure*}
\begin{figure*}[t]
    \begin{center}
        \includegraphics[trim=0.1cm 0.2cm 0.1cm 0.3cm,clip,width=1.0\linewidth,keepaspectratio]{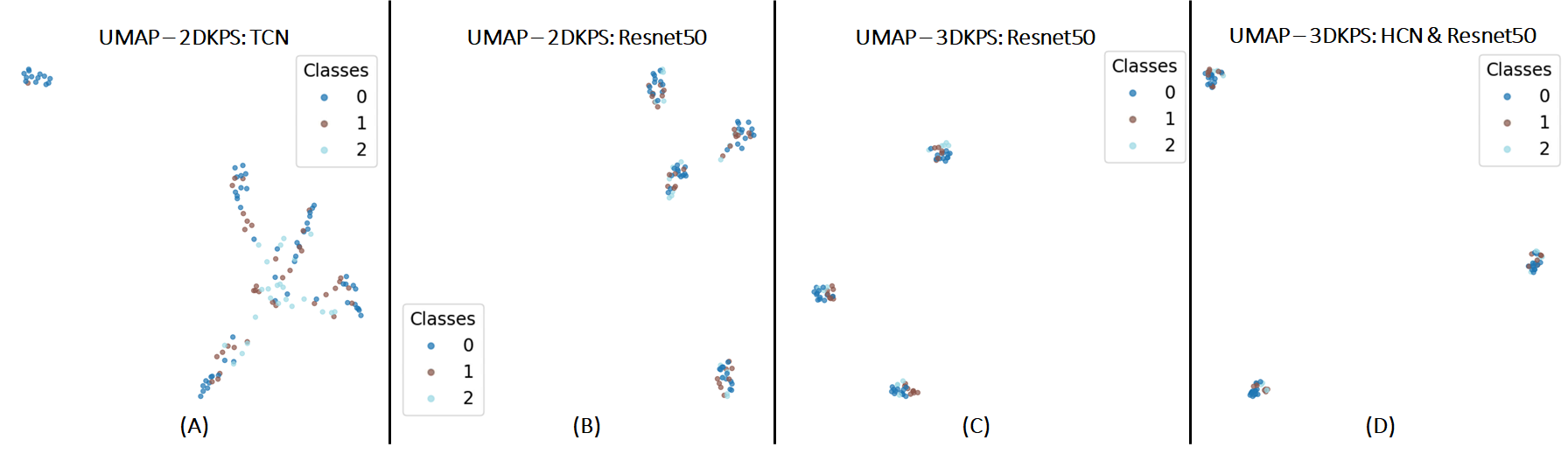}
        \caption{UMAP embedding clusters based on the representative features for PIGD classification on test data for different models and input data (A) TCN based on 2D joints (B) Resnet50 based on 2D joints (C) Resnet50 based on 3D joints (D) Ensemble of HCN + Resnet50 based on 3D joints}
        \label{fig_umap_pigd_net}
    \end{center}
\end{figure*}
\begin{figure*}[t]
    \begin{center}
        \includegraphics[trim=0.1cm 0.2cm 0.1cm 0.3cm,clip,width=0.95\linewidth,keepaspectratio]{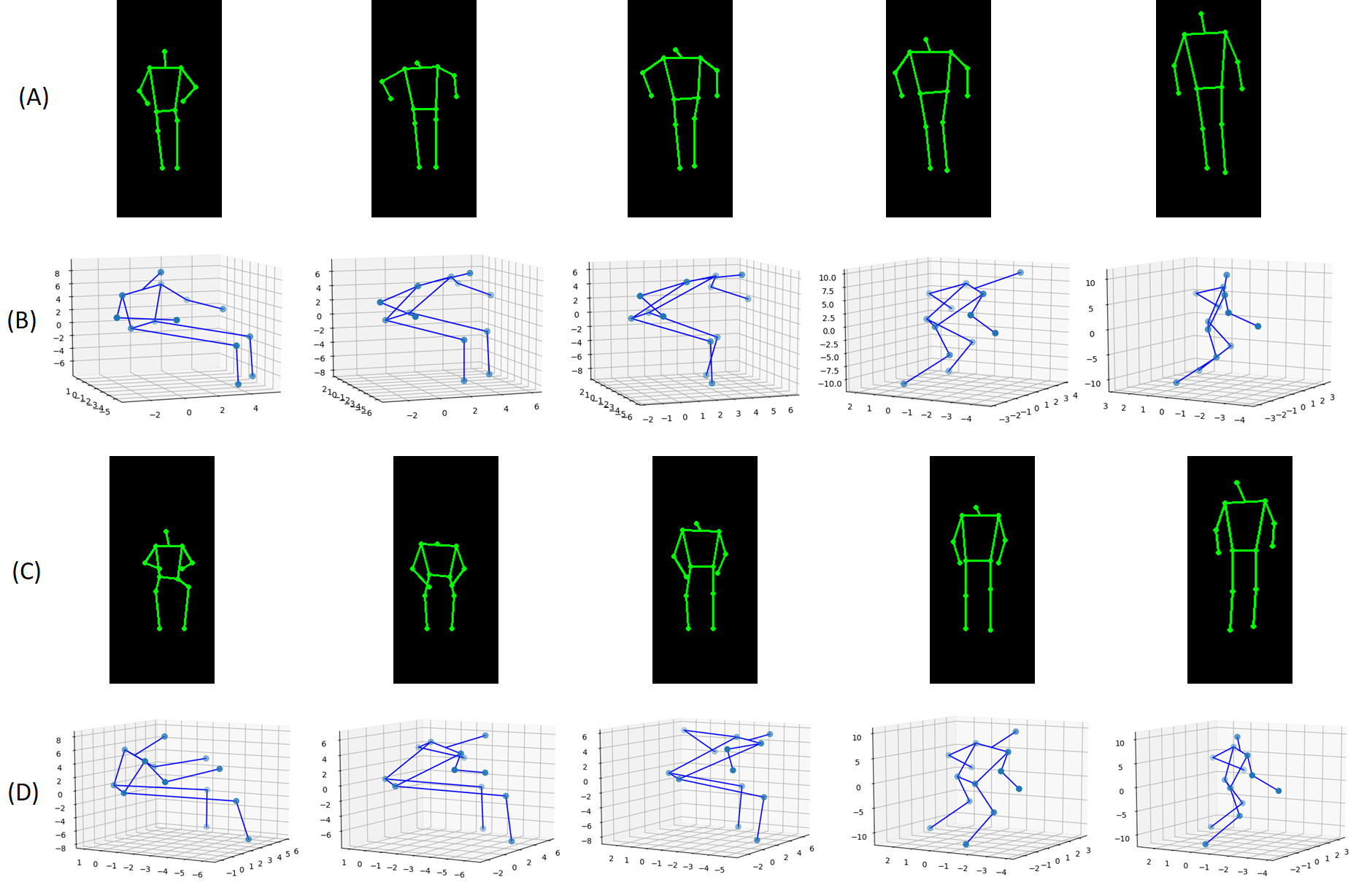}
        \caption{Samples of sit-stand videos from our dataset where the patient is firstly in sitting position and then stands. (A) and (C) relate to the 2D poses extracted from such sequences for two different video inputs; (B) and (D) represent the corresponding 3D poses for the video sequence (represented with an angle view change in the \textit{xy} dimension for better viewing). Note that it is difficult to classify video sequences based only on the 2D pose as there are very little changes in the pose information compared to the 3D pose.}
        \label{fig_2d3dpose}
    \end{center}
\end{figure*}
\subsection{Results}
In this section, we present the results of our baseline models, ensemble models and benchmark them with those of the neurologist video raters for the task of predicting the UPDRS scores. For benchmarking purposes, we convert the raw BRADY and PIGD scores provided by the clinician video raters into integers.
\subsubsection{Baseline Models}
\begin{table}[h]
	\caption{Comparisons of the weighted f1 scores for models trained using the different input keypoints - 2D \& 3D body joints respectively for BRADY and PIGD classification with 5-fold stratified split protocol.}
	\vspace{5pt}
	\label{table_baseline_f1}
	\centering
	\setlength\tabcolsep{5.0pt}
	\begin{tabular}{@{}lcccccccc@{}}
		\toprule
		Model & Input & BRADY & PIGD\\
		\midrule
		MLP & Body 3D joints &0.36&0.37\\
		LSTM & Body 3D joints&0.28&0.35\\
		TCN & Body 3D joints&0.45&0.58\\
		HCN & Body 3D joints&0.47&0.68\\
		Resnet50 & Body 3D joints&0.73&0.72\\
		ST-GCN & Body 3D joints&0.62&0.70\\
		\hline
		MLP & Body 2D joints &0.35&0.33\\
		LSTM & Body 2D joints &0.17&0.20\\
		TCN & Body 2D joints &0.41&0.46\\
		HCN & Body 2D joints &0.43&0.53\\
		Resnet50 & Body 2D joints &0.65&0.64\\
	    ST-GCN & Body 2D joints &0.57&0.54\\
		\bottomrule
	\end{tabular}
	\vspace{5pt}
\end{table}
As we have an imbalanced class distribution for both BRADY and PIGD, we employ the metric of weighted \textit{f1 score} to benchmark our various models. We compare all our baseline models mentioned earlier in Section 3 on both 2D joints and 3D joints. The results for the performance are shown in Table \ref{table_baseline_f1}. It can be seen that very simple models such as MLP, LSTM, and TCN are not able to achieve even a reasonable score for these tasks. As the models become more complex in nature, they are able to learn the relevant features and as such HCN, ST-GCN, and Resnet50 perform much better. Another important observation from the baseline performance results is that for all the models the \textit{f1 scores} for 3D joints are comparatively higher than those for 2D joints. To analyze the results of the models in more detail, we extract the representative feature embeddings (256 dimensional) of the models from the penultimate layer of each model on our test dataset. We apply UMAP technique to map out the representative features which are shown in Fig \ref{fig_umap_brady_net} and Fig \ref{fig_umap_pigd_net} for the tasks of BRADY and PIGD classification respectively. It can be seen that for Resnet50 the learned features are more clustered in nature compared to that of TCN where they are quite scattered. However, for both the networks, the clustering is much better compared to the raw 2D and 3D joints which was shown in Fig \ref{fig_umap_all}. Moreover, it can also be noted that the variance amongst the clusters in case of Resnet50 is less for 3D joints as compared to the 2D joints, thus explaining the higher performance of our models in Table \ref{table_baseline_f1} for 3D joints compared to the 2D joints. We also show some qualitative results of 2D pose and 3D pose sequences to better understand the comparatively higher performance for 3D joints in Fig \ref{fig_2d3dpose}. It can be seen that the 3D pose information is much more distinctive for these kind of sit-stand motion especially to identify the nuanced features and changes between the bones responsible for variation in BRADY and PIGD scores.
\begin{figure*}[h]
    \begin{center}
        \includegraphics[trim=0.1cm 0.2cm 0.1cm 0.3cm,clip,width=1.0\linewidth,keepaspectratio]{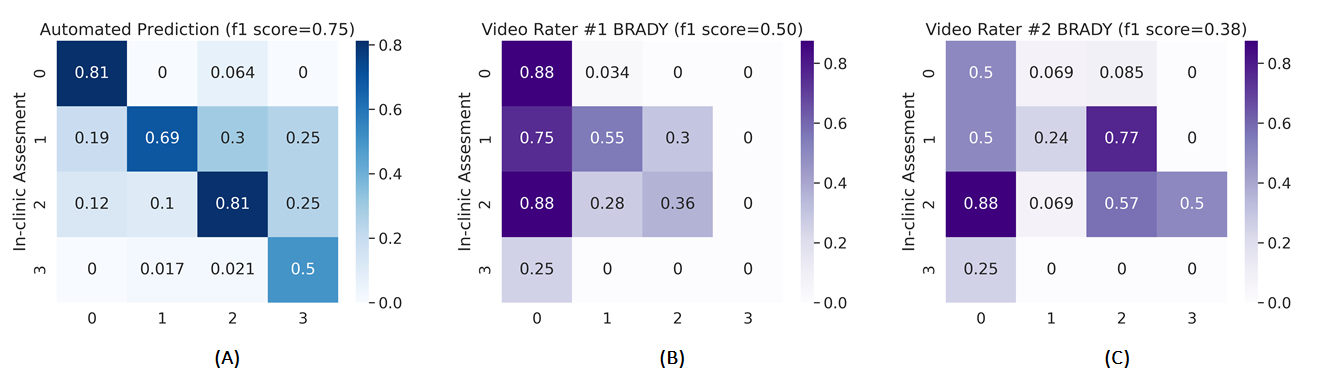}
        \caption{Comparison of confusion matrices for the task of BRADY for 5-fold split protocol (A) Our ensemble model of ST-GCN + Resnet50 based on 3D joints input, (B) Clinician video rater \#1 based on video input, (C) Clinician video rater \#2 based on video input}
        \label{fig_brady_vr_split}
    \end{center}
\end{figure*}
\begin{figure*}[h]
    \begin{center}
        \includegraphics[trim=0.1cm 0.2cm 0.1cm 0.3cm,clip,width=1.0\linewidth,keepaspectratio]{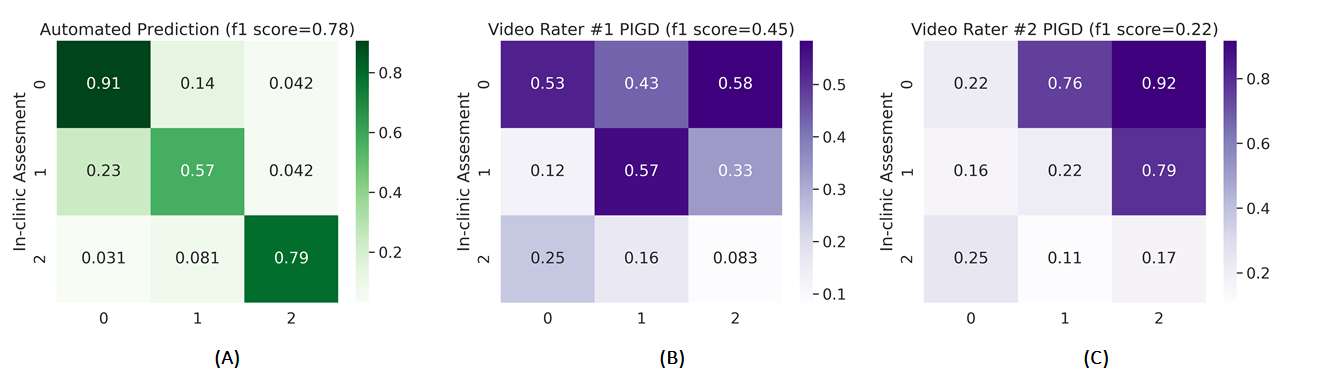}
        \caption{Comparison of confusion matrices for the task of PIGD for 5-fold split protocol (A) Our ensemble model of HCN + Resnet50 based on 3D joints input, (B) Clinician video rater \#1 based on video input, (C) Clinician video rater \#2 based on video input}
        \label{fig_pigd_vr_split}
    \end{center}
\end{figure*}
\subsubsection{Ensemble Models}
\begin{table}[h]
	\caption{Comparisons of the weighted f1 scores for ensemble models trained using the 3D joints with clinician video raters for BRADY and PIGD classification with 5-fold stratified split protocol.}
	\vspace{5pt}
	\label{table_ensemble_f1}
	\centering
	\setlength\tabcolsep{5.0pt}
	\begin{tabular}{@{}lcccccccc@{}}
		\toprule
		Model & Input & BRADY & PIGD\\
		\midrule
		Clinician Video Rater \#1 & Video &0.50&0.45\\
		Clinician Video Rater \#2 & Video &0.38&0.22\\
		\hline
		\textbf{(Ours) ST-GCN + Resnet50 } & 3D joints &\textbf{0.75}&0.70\\
		\textbf{(Ours) HCN + Resnet50} & 3D joints &0.72&\textbf{0.78}\\
		\textbf{(Ours) ST-GCN + HCN} & 3D joints &0.65&0.74\\
		\bottomrule
	\end{tabular}
\end{table}
Having performed the initial baseline experiments with different models for 2D and 3D joints, we build ensemble models combining the higher performing models from the baseline experiments. For this purpose, we resort to three models - ST-GCN, HCN and Resnet50. We build the ensembles for each of their combination i.e. ST-GCN \& HCN, ST-GCN \& Resnet50, and HCN \& Resnet50 based on the input of 3D joints as the performance for 3D joints is relatively higher compared to those based on 2D joints. The performance results for our ensemble models are shown in Table \ref{table_ensemble_f1} where we also compare it with those of the clinician video raters. It can be seen from the results that our models surpass the scores of the clinician video raters for both the tasks of BRADY and PIGD. More specifically, the highest f1 score obtained by our models is \textbf{0.75} and \textbf{0.78} for BRADY and PIGD respectively compared to those of the clinician video raters which is \textbf{0.50} and \textbf{0.45}. Note that we also train a \textit{combined ensemble} of ST-GCN, HCN, and Resnet50, however it's performance is on par with their individual respective combinations, so we do not present it here.
We also present confusion matrices for the tasks of BRADY and PIGD for our best performing ensemble and the clinician video raters in Fig \ref{fig_brady_vr_split} and Fig \ref{fig_pigd_vr_split} respectively. It can be seen that our ensemble models performance is higher for each individual class of BRADY and PIGD except class \{0\} for BRADY when compared to those of the clinicians who have provided the ratings based on the video input. The better performance of the ensemble models can be also be justified by noting the tight clustering of the learned features by ensemble models shown in Fig \ref{fig_umap_brady_net} and Fig \ref{fig_umap_pigd_net}.

\section{Conclusion}
In this paper, we explore the feasibility of automatically assessing the motor symptoms of Parkinson's disease patients from video. To this end, we collected video clips containing sit-stand motions from 35 subjects, during two separate clinic visits. This task was performed as part of wider UPDRS tests under the supervision of a neurologist who also assigned scores for each task. The scores range from zero, for no impairment, to four, corresponding to maximum impairment. At the end of the session, all the individual scores were added up to form the total UPDRS score. This measure serves an essential clinical role in assessing disease progression as well as customizing treatment for individual patients to improve their specific motor symptoms. Here we focused on bradykinesia (BRADY) and posture and gait disorders (PIGD) and used the corresponding sub-scores from UPDRS as end-points for deep learning models. We then demonstrate that it is possible to predict BRADY and PIGD scores from just a short sit-stand video clip. Specifically, the F1-scores of our models for BRADY and PIGD end-points were $0.75$ and $0.78$, respectively, outperforming the best results from two clinician video raters ($0.50$ and $0.45$) benchmarked against in-clinic assessments. These results suggest the presented framework has potential to provide continual clinically acceptable end-points of PD without imposing additional burden on clinicians and patients. On further validation, automatic video analysis could unlock an array of use cases such as enhanced tele-medicine or clinical-grade at-home assessments.

\FloatBarrier
{\small
\bibliographystyle{ieee_fullname}
\bibliography{cvpr}
}

\end{document}